\documentclass[lettersize,journal]{IEEEtran}
\usepackage{amsmath,amsfonts}
\usepackage{algorithmic}
\usepackage{algorithm}
\usepackage{array}
\usepackage[caption=false,font=normalsize,labelfont=sf,textfont=sf]{subfig}
\usepackage{textcomp}
\usepackage{stfloats}
\usepackage{url}
\usepackage{verbatim}
\usepackage{graphicx}
\usepackage{cite}
\usepackage{amssymb}
\usepackage{multirow}
\usepackage[table]{xcolor}
\usepackage{pifont}

\newcommand{\cmark}{\ding{51}}
\newcommand{\xmark}{\ding{55}}

\newcommand{\bluecheck}{\raisebox{-0.2em}{\includegraphics[height=1.2em]{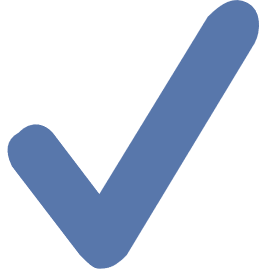}}}
\newcommand{\greencheck}{\raisebox{-0.2em}{\includegraphics[height=1.2em]{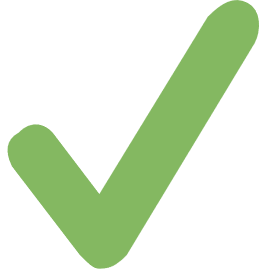}}}
\newcommand{\target}{\raisebox{-0.2em}{\includegraphics[height=1.2em]{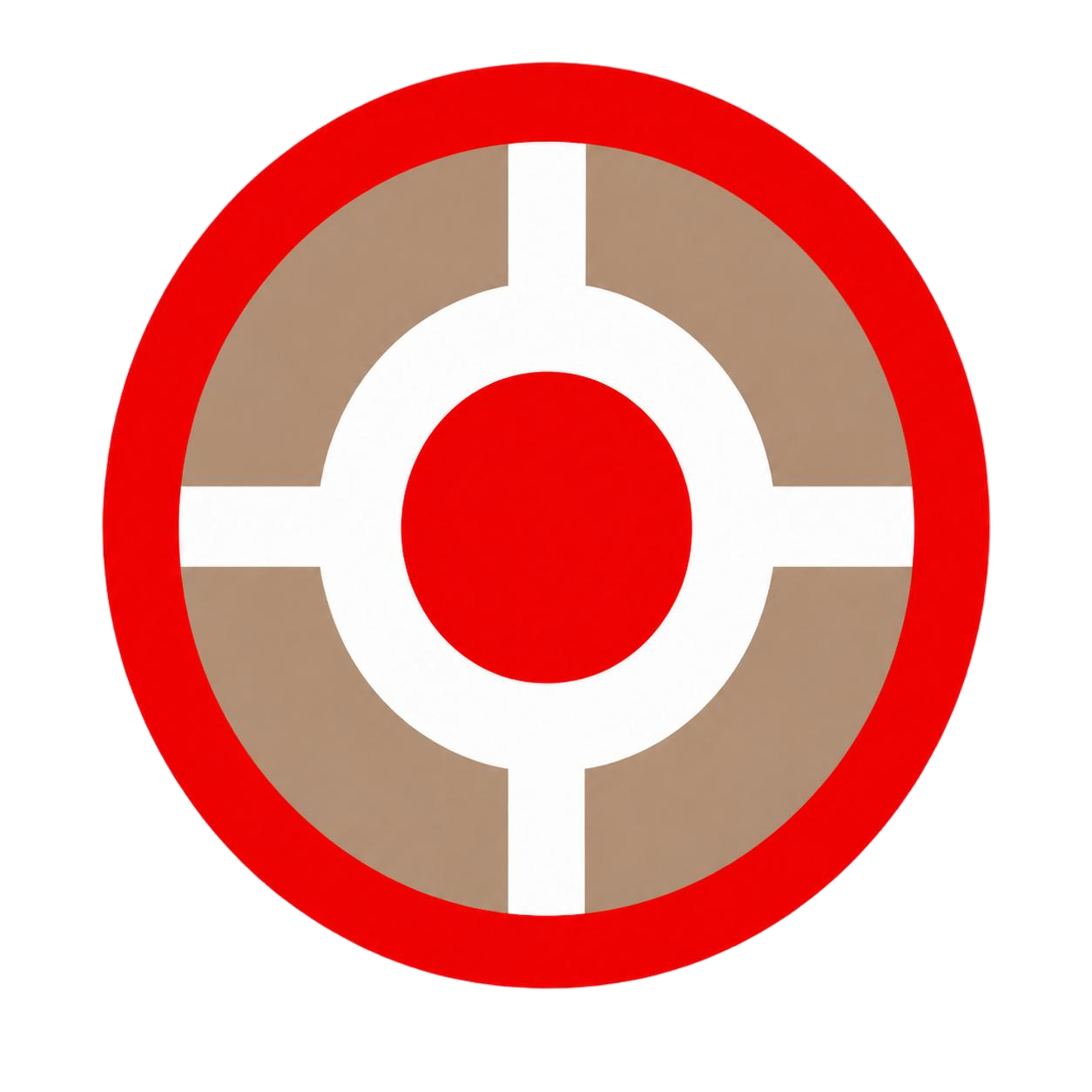}}}

\definecolor{mygreen}{RGB}{34,139,34}
\definecolor{myred}{RGB}{180,40,40}

\begin{document}

\title{P2DNav: Panorama-to-Downview Reasoning for Zero-shot Vision-and-Language Navigation}

\author{Kai Sheng, Liuyi Wang, Haojie Dai, Jinlong Li, Yongrui Qin, Zongtao He, \\Chengju Liu and Qijun Chen$^\dagger$, ~\IEEEmembership{Senior Member,~IEEE}
\thanks{This paper is supported by the National Natural Science Foundation of China under Grants (62233013 and 62473295). ($^\dagger$Corresponding author: Qijun Chen.)}

\thanks{The authors are with the Department of Control Science and Engineering, Tongji University, Shanghai 210804, China. E-mail: 2254293@tongji.edu.cn, wly@tongji.edu.cn, tju\_dhj@tongji.edu.cn, li\_jinlong@tongji.edu.cn, 2311441@tongji.edu.cn, zongtao.he@keenon.com, liuchengju@tongji.edu.cn, qjchen@tongji.edu.cn.}}

\markboth{Journal of \LaTeX\ Class Files,~Vol.~14, No.~8, August~2021}%
{K. Sheng \MakeLowercase{\textit{et al.}}: DREAM: Dynamic Routing of Experts via Attention-based Mixture for Vision-Language-Action Modeling}

\IEEEpubid{0000--0000/00\$00.00~\copyright~2021 IEEE}

\maketitle

\begin{abstract}
Vision-and-language navigation (VLN) requires an embodied agent to ground natural-language instructions into executable navigation actions in unseen environments. Existing zero-shot methods typically rely on additional waypoint prediction modules, which often entangle high-level directional reasoning with fine-grained local grounding, leading to error-prone and unstable decisions. In this paper, we propose P2DNav, a hierarchical framework for zero-shot vision-and-language navigation. P2DNav consists of three core components: Panorama-to-Downview (P2D), Sliding-Window Dialogue Memory (SDM), and Reflective Reorientation Mechanism (RRM). P2D explicitly decomposes navigation decision-making into two stages: panoramic direction selection and downview local grounding. It first selects the instruction-relevant direction from a 360° panorama, and then predicts a pixel-level target point from the downview RGB observation in that direction. In addition, SDM organizes navigation history as a multi-turn dialogue context and maintains recent visual observations within a sliding window to support long-horizon navigation. RRM further enables reflective reorientation by assessing the reliability of local grounding based on the downview observation and returning to panoramic direction selection when necessary. Experiments on the R2R-CE benchmark show that P2DNav achieves strong performance among zero-shot methods. In particular, compared with the state-of-the-art (SOTA) zero-shot waypoint-based and waypoint-free methods, P2DNav achieves SR gains of 146.6\% and 58.9\%, respectively, demonstrating the effectiveness of P2D, SDM, and RRM for zero-shot VLN. Code will be released for public use.
\end{abstract}


\section{Introduction}
\label{intro}
\IEEEPARstart{V}{ision-and-language} navigation (VLN) requires an embodied agent to follow natural-language instructions and navigate in previously unseen environments~\cite{anderson2018vision}.
To solve this task, the agent must ground language into visual observations, infer the intended direction of movement, and execute feasible navigation actions~\cite{latent}.
Humans often navigate unfamiliar environments in a hierarchical manner: they first look around to determine the overall direction, and then look down along that direction to decide where to step next.
This intuitive process reveals a key structure of VLN: high-level directional reasoning determines \emph{which direction to go}, while low-level local grounding determines \emph{where exactly to move}.

Recent zero-shot VLN methods~\cite{qiao2025open,zhou2024navgpt,long2024instructnav} typically leverage large language models (LLMs) and vision-language models (VLMs)~\cite{achiam2023gpt,yang2023dawn,bai2025qwen3} for semantic reasoning and often rely on additional waypoint prediction modules for executable navigation~\cite{waypoint,vlnbert,mossvln}.
However, they do not explicitly model this hierarchical structure, and instead couple directional reasoning with local grounding within a single waypoint or action decision.
This coupling makes it difficult to isolate and revise errors from different decision levels, since directional ambiguity and local target inaccuracy are handled together within a single step.
This issue becomes more pronounced in visually complex continuous environments~\cite{beyond}, where an ambiguous direction choice or an inaccurate local target may directly lead to navigation deviation.

\begin{figure}[t]
    \centering
    \includegraphics[width=0.5\textwidth]{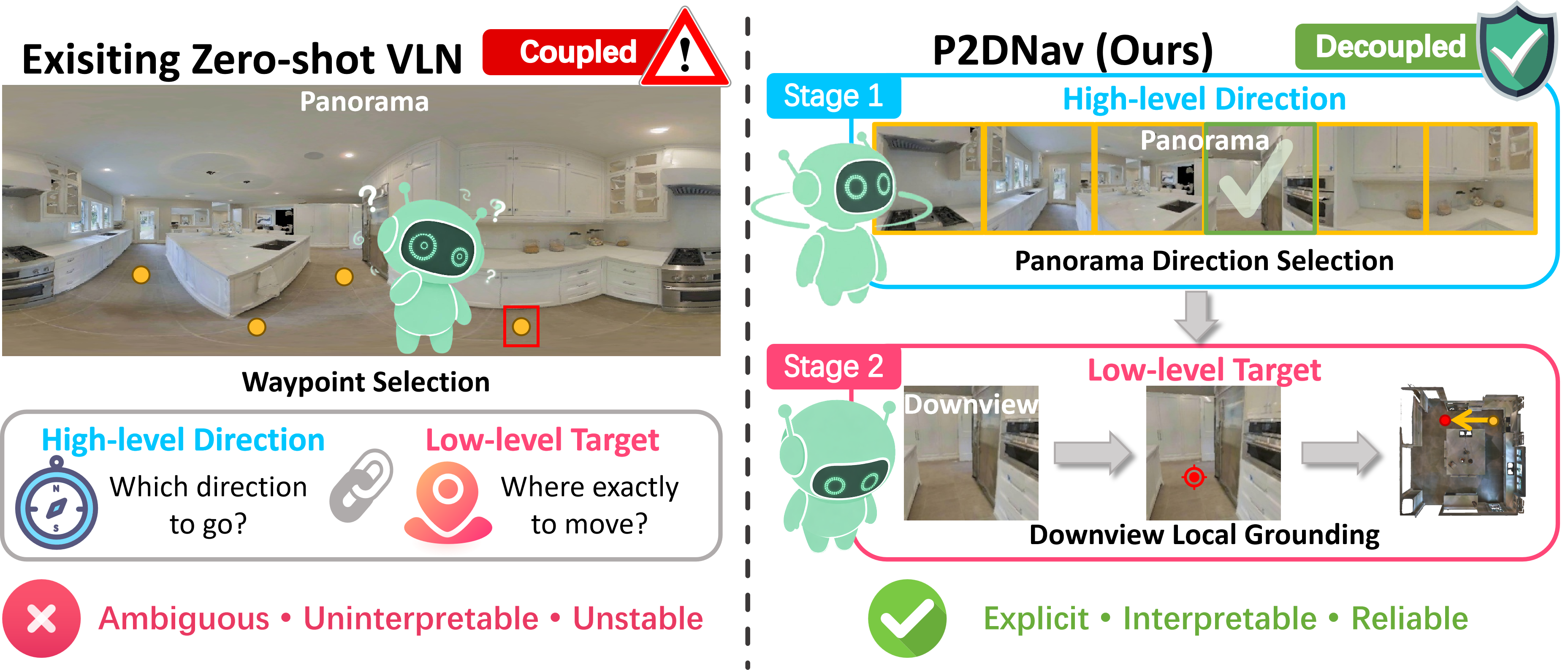}
    \caption{Comparison between existing zero-shot VLN methods (e.g., Open-Nav~\cite{qiao2025open}, NavGPT~\cite{zhou2024navgpt}, and SmartWay~\cite{shi2025smartway}) and P2DNav. Existing methods often couple high-level directional reasoning and low-level local grounding within a single waypoint decision. P2DNav explicitly decomposes this process into two sequential stages: panoramic direction selection for coarse movement reasoning and downview local grounding for precise target prediction.}
    \label{fig:head}
\end{figure}

\IEEEpubidadjcol

Motivated by this observation, we propose P2DNav, a hierarchical zero-shot VLN framework (see Fig.~\ref{fig:head}). 
P2DNav consists of three core modules: Panorama-to-Downview (P2D), Sliding-Window Dialogue Memory (SDM), and Reflective Reorientation Mechanism (RRM).
P2D serves as the core navigation paradigm of P2DNav and follows a coarse-to-fine process. 
It first selects an instruction-relevant directional subview from a stitched 360$^\circ$ panorama, enabling global reasoning over candidate movement directions. 
Given the selected direction, it then uses the corresponding downview RGB observation to predict a pixel-level local target for executable navigation. 
Compared with panoramic observations, the downview observation emphasizes nearby traversable regions and local geometric cues, making it more suitable for fine-grained grounding. 
By organizing direction selection and local target prediction in sequence, P2D decouples high-level directional reasoning from low-level local grounding, without relying on an additionally trained waypoint generator.

Beyond the current-step hierarchy provided by P2D, navigation decisions also require long-horizon memory, since past observations, selected directions, and intermediate grounding results provide useful context for later steps. 
Existing zero-shot methods often maintain such history with additional memory summarization modules, which introduce extra overhead and may repeatedly compress or overwrite fine-grained details~\cite{shi2025smartway}. 
To address this issue, we introduce Sliding-Window Dialogue Memory (SDM), which represents navigation history as a structured multi-turn dialogue rather than a repeatedly rewritten summary.
SDM preserves textual instructions, model responses, and decision records as persistent context, while applying a sliding window to visual observations.
Since visual inputs dominate context length compared with text, retaining only recent visual turns prevents unbounded context growth while keeping the visual evidence most relevant to local decision-making.
In this way, SDM provides efficient, multimodal, and comprehensive memory support for navigation.

Although P2D decouples direction selection from local grounding, the downview stage still needs to assess whether the selected view can support reliable target prediction.
To this end, we introduce Reflective Reorientation Mechanism (RRM), which evaluates the grounding reliability of the selected downview observation before execution.
When the observation lacks sufficient traversability cues or instruction-relevant evidence, RRM rejects the current local grounding and returns the decision process to panoramic direction selection with failure feedback.
This allows P2DNav to reconsider the previous directional choice and select a more reliable subview.
Consequently, RRM forms a reflective loop between direction selection and local grounding, preventing uncertain downview predictions from being directly executed.

Experiments on the R2R-CE benchmark demonstrate the effectiveness of P2DNav under the training-free setting.
P2DNav achieves strong performance among zero-shot methods and surpasses the state-of-the-art (SOTA) waypoint-based and waypoint-free methods by 146.6\% and 58.9\% in SR, respectively.
Further analysis shows that the proposed P2D, SDM, and RRM contribute to more stable navigation by decomposing directional reasoning and local grounding, preserving useful historical context, and enabling reflective reorientation.

Our contributions are summarized as follows:
\begin{itemize}
    \item We propose P2DNav, a hierarchical zero-shot VLN framework that integrates P2D, SDM, and RRM for training-free navigation in continuous environments.
    \item We introduce Panorama-to-Downview (P2D) decision, which decouples high-level directional reasoning from low-level local grounding.
    \item We introduce Sliding-Window Dialogue Memory (SDM), which preserves multi-turn navigation history with a sliding window for efficient long-horizon memory.
    \item We introduce Reflective Reorientation Mechanism (RRM), which assesses the reliability of downview local grounding and returns unreliable cases to panoramic direction selection for reorientation.
    \item Extensive experiments on R2R-CE demonstrate that P2DNav achieves state-of-the-art (SOTA) performance among zero-shot methods and validates the effectiveness of P2D, SDM, and RRM in zero-shot VLN.
\end{itemize}

\section{Related Work}
\label{relate}

\subsection{Vision-and-Language Navigation}

Vision-and-Language Navigation (VLN) requires an embodied agent to follow natural-language instructions and visual observations to reach a target location~\cite{anderson2018vision}. Early VLN methods were mainly developed in discrete environments, where agents navigate over predefined connectivity graphs and select among candidate viewpoints~\cite{wang2024vision,flexvln}. Although this formulation supports high-level language grounding and route planning, it abstracts away fine-grained low-level movements and is therefore less realistic for embodied navigation.
With the introduction of continuous-environment benchmarks~\cite{beyond,wang2025rethinking}, agents are required to directly execute low-level actions in 3D environments. To bridge language understanding and continuous control, existing methods typically learn end-to-end action policies or rely on waypoint-based supervision~\cite{chen2021history}. Subsequent studies further improve navigation performance through stronger visual-language representation learning, reinforcement learning, and data augmentation~\cite{mee,he2025multilevel}. In parallel, many methods explicitly construct spatial representations or introduce collaborative decision-making mechanisms, such as topological maps in DUET~\cite{chen2022duet} and ETPNav~\cite{etpnav}, bird's-eye-view representations in BEVBert~\cite{an2022bevbert}, and the hybrid large-small model collaborative decision system in CLASH~\cite{wang2025clash}.
These methods have significantly advanced VLN by improving spatial reasoning and long-horizon navigation. However, most of them still depend on task-specific training, supervised waypoint prediction, or learned spatial representations. In contrast, our work explores zero-shot continuous VLN, aiming to bridge high-level semantic intent and low-level executable grounding without learned waypoint proposal modules or task-specific action policies.

\subsection{Zero-shot VLN with Foundation Models}

Foundation models, including large language models (LLMs) and vision-language models (VLMs), are pretrained on broad-scale data and can be adapted or prompted for diverse downstream tasks~\cite{achiam2023gpt,yang2023dawn,bai2025qwen3}. Their capabilities in semantic reasoning, instruction following, task planning, and multimodal understanding make them promising for embodied navigation. Recently, zero-shot VLN methods have explored using foundation models as navigators. In discrete environments, DiscussNav, NavGPT, and MapGPT leverage visual descriptions, navigation history, or topological maps as prompts for step-by-step decision making~\cite{long2023discuss,zhou2024navgpt,chen2024mapgpt}. These methods demonstrate the potential of pretrained models for zero-shot VLN, but mainly focus on high-level navigation decisions in discrete settings.
Recent works further extend zero-shot navigation to continuous environments. Open-Nav explores zero-shot VLN-CE with open-source foundation models~\cite{qiao2025open}. InstructNav unifies diverse instruction-navigation tasks via Dynamic Chain-of-Navigation and Multi-sourced Value Maps~\cite{long2024instructnav}. CA-Nav introduces constraint-aware sub-instruction switching based on egocentric observations~\cite{CANav}, while SmartWay combines enhanced waypoint prediction with history-aware reasoning and adaptive backtracking~\cite{shi2025smartway}. Although these methods reduce the need for task-specific navigation training, many of them still rely on additional waypoint prediction modules, value maps, or topological structures. Moreover, their decision processes often couple high-level directional reasoning with low-level executable grounding, while navigation memory often requires repeated updating or summarization, introducing extra overhead and risking the loss of fine-grained historical details.
Motivated by these limitations, our work does not simply prompt foundation models to make one-step waypoint or action decisions. Instead, we explicitly decouple high-level directional reasoning from low-level executable grounding, reducing the instability caused by coupled navigation decisions. Meanwhile, we organize multi-turn navigation history with a sliding-window memory mechanism, providing comprehensive yet efficient context for long-horizon decision making.

\begin{figure*}[t]
    \centering
    \includegraphics[width=1\textwidth]{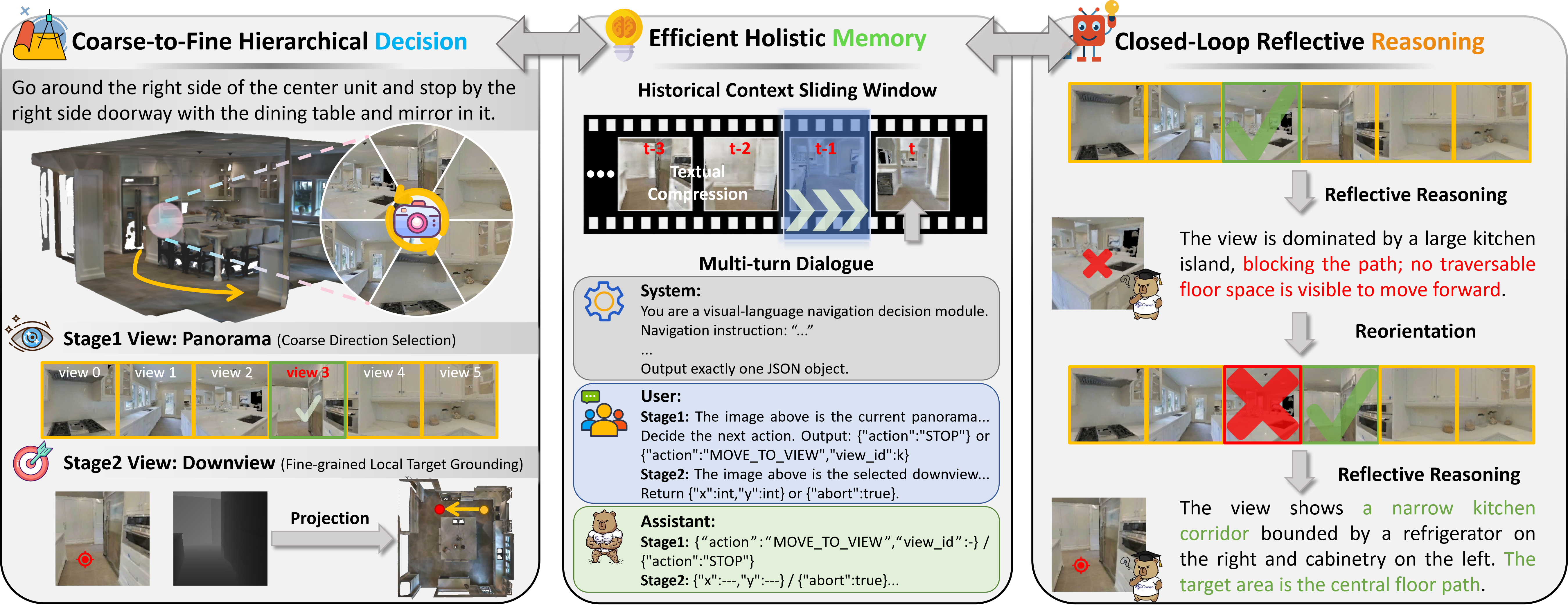}
    \caption{Overview of P2DNav. From left to right, the framework illustrates three components: (1) Panorama-to-Downview Hierarchical Decision (P2D), which decomposes navigation into panoramic directional selection and downview local grounding; (2) Sliding-Window Dialogue Memory (SDM), which maintains efficient multi-turn historical context with a sliding-window strategy; and (3) Reflective Reorientation Mechanism (RRM), which enables the agent to return from unreliable local grounding to panoramic reorientation.}
    \label{fig:p2d}
\end{figure*}

\section{Method}
\label{method}

\subsection{Overview}

We propose P2DNav, a zero-shot VLN framework that reformulates navigation as a hierarchical, memory-aware, and reflective decision process rather than a one-pass action prediction problem.
As illustrated in Fig.~\ref{fig:p2d}, P2DNav consists of three core components: Panorama-to-Downview (P2D) hierarchical decision-making, Sliding-Window Dialogue Memory (SDM), and Reflective Reorientation Mechanism (RRM).

The remainder of this section is organized as follows.
Sec.~\ref{sec:problem_formulation} formulates zero-shot VLN as a hierarchical decision process.
Sec.~\ref{sec:p2d} introduces P2D, which decouples panoramic direction selection from downview local grounding.
Sec.~\ref{sec:sdm} presents SDM, which provides efficient long-horizon memory through structured multi-turn dialogue.
Sec.~\ref{sec:rrm} describes RRM, which enables reliability-aware reorientation when local grounding is unreliable.

\subsection{Problem Formulation}
\label{sec:problem_formulation}

At navigation step $t$, the agent is given a language instruction $I$ and the current visual observations in a continuous environment.
Instead of directly mapping the observation-instruction pair to a one-step target, we formulate zero-shot VLN as a hierarchical decision process with memory conditioning and reflective reorientation.
The agent first performs coarse direction selection from a panoramic observation, and then conducts fine-grained local grounding using only the downview observation aligned with the selected direction.

Formally, let $P_t$ denote the panoramic observation at step $t$, $k_t$ denote the selected direction, $D_t^{k_t}$ denote the downview observation associated with this selected direction, and $\mathcal{M}_t$ denote the maintained dialogue memory.
The overall decision process is formulated as
\begin{equation}
\tau_t = \mathcal{F}(I, P_t, D_t^{k_t}, \mathcal{M}_t),
\end{equation}
where $\tau_t$ denotes the final navigation decision at step $t$, such as stopping or producing a local navigation target.
Here, $\mathcal{F}$ denotes the overall prompting-based decision procedure.
In P2DNav, $\mathcal{F}$ is instantiated through three complementary components.
P2D implements the current-step hierarchical decision process by first selecting $k_t$ from $P_t$ and then grounding a local target from $D_t^{k_t}$, SDM maintains dialogue memory $\mathcal{M}_t$ for historical context, and RRM redirects unreliable local grounding back to panoramic direction selection for reorientation.
We begin with the P2D hierarchy.

\subsection{Panorama-to-Downview Hierarchical Decision (P2D)}
\label{sec:p2d}

Following the above formulation, P2D instantiates the hierarchical part of the decision process.
It decomposes each navigation decision into two sequential stages: \emph{panoramic direction selection} and \emph{downview local grounding}.
The first stage reasons over the surrounding environment at a coarse directional level, while the second stage focuses on fine-grained target prediction within the selected direction.
This design allows the model to separate the question of \emph{which direction to go} from \emph{where exactly to move}.

At step $t$, the agent observes $K$ directional RGB views from its current pose and constructs a stitched panoramic observation:
\begin{equation}
P_t = \mathrm{Stitch}(v_t^0, v_t^1, \dots, v_t^{K-1}),
\end{equation}
where $v_t^k$ denotes the RGB subview corresponding to direction $k$, and the choice of $K$ is analyzed in Sec.~\ref{sec:ablation}.
The $K$ directional subviews are arranged into a single panoramic image according to their spatial order.
We further annotate each subview with its view index on the top of the panorama, providing explicit visual prompts that help the model associate image regions with selectable directions.
This panoramic representation exposes all candidate movement directions in a unified visual context, enabling the model to compare them according to the instruction and navigation history.

Given the instruction $I$, the panorama $P_t$, and the dialogue memory $\mathcal{M}_t$, the first stage queries the VLM for a coarse navigation decision:
\begin{equation}
d_t = Q_{\mathrm{P}}(I, P_t, \mathcal{M}_t).
\end{equation}
The decision $d_t$ is defined over the following output space:
\begin{equation}
d_t \in \{\texttt{STOP}\} \cup \{\texttt{MOVE\_TO\_VIEW}(k)\}_{k=0}^{K-1}.
\end{equation}
Here, $Q_{\mathrm{P}}$ denotes a zero-shot panoramic query to the VLM for direction selection.
If the current location already satisfies the instruction, the model outputs \texttt{STOP}.
Otherwise, it selects the directional subview that is most relevant to the instruction, denoted as $d_t=\texttt{MOVE\_TO\_VIEW}(k_t)$, where $k_t$ is the selected direction index.

After a direction $k_t$ is selected, P2D switches from panoramic reasoning to downview local grounding.
The agent obtains a direction-aligned downview RGB observation $D_t^{k_t}$ for the selected direction.
Compared with panoramic observations, the downview emphasizes nearby traversable regions and local geometric cues, making it more suitable for precise target prediction.

The second stage queries the VLM to predict a pixel-level local target from the selected downview:
\begin{equation}
p_t = Q_{\mathrm{D}}(I, D_t^{k_t}, \mathcal{M}_t), \qquad p_t=(u_t,v_t).
\end{equation}
Here, $Q_{\mathrm{D}}$ denotes a zero-shot downview query to the VLM for local grounding.
The predicted pixel target is then projected to the world coordinate system using the aligned depth observation and camera geometry, and the resulting point is used as a short-horizon target for the low-level controller.

Through this coarse-to-fine process, P2D decouples high-level directional reasoning from low-level local grounding without relying on a separately trained waypoint generator.
In our implementation, both $Q_{\mathrm{P}}$ and $Q_{\mathrm{D}}$ are instantiated by Qwen-VL-series VLMs through zero-shot prompting~\cite{bai2025qwen3}.
However, each decision is still part of a longer navigation episode, where previous observations and decisions can provide useful context for later steps.
We therefore incorporate memory directly into the interaction process through SDM.

\subsection{Sliding-Window Dialogue Memory (SDM)}
\label{sec:sdm}

Sliding-Window Dialogue Memory (SDM) maintains such cross-step context by organizing navigation history as a structured multi-turn dialogue.
Many prior foundation-model-based navigation methods~\cite{qiao2025open,zhou2024navgpt,long2024instructnav} still operate largely in a single-turn manner.
Even when historical information is used, it is often introduced through an additional memory summarization stage after individual reasoning steps, which increases complexity and may discard useful interaction structure.
Instead of treating each navigation step as an isolated prompt, SDM formulates navigation history as a native multi-turn dialogue process.
Each episode is organized as a sequence of \texttt{system}, \texttt{user}, and \texttt{assistant} turns.
The \texttt{system} turn specifies the navigation role, the two-stage protocol, and the output format.
The \texttt{user} turns provide the current visual observation together with stage-specific questions, while the \texttt{assistant} turns record the model's structured decisions.
In this way, the model reasons from both the current observation and a structured history of previous observations and decisions.

We denote the accumulated dialogue history up to step $t$ as
\begin{equation}
\mathcal{H}_t = \{U_1,A_1,U_2,A_2,\dots,U_t,A_t\},
\end{equation}
where $U_i$ and $A_i$ denote the user and assistant turns at step $i$, respectively.
Preserving the full history would make the context grow rapidly over long episodes, especially because visual inputs consume much more context than text and introduce substantial computational overhead.
Therefore, SDM introduces a sliding-window mechanism:
\begin{equation}
\mathcal{M}_t = \mathrm{Window}(\mathcal{H}_t; W),
\end{equation}
where the most recent $W$ steps retain their full visual content, while images outside the sliding window are discarded and only their text-form decision history is preserved.
This design preserves recent multimodal evidence while preventing the visual context from growing unbounded over long episodes.

SDM is not a separate external summarizer, but a memory organization strategy embedded in the interaction process itself.
It keeps the decision history in a structured dialogue form and selectively retains recent visual observations for subsequent reasoning.
Overall, SDM complements P2D by allowing each hierarchical decision to condition on previous observations and decisions.
Nevertheless, historical context alone cannot guarantee that the selected downview provides reliable evidence for local grounding.
P2DNav therefore needs a mechanism to detect such unreliable cases before execution and reconsider the directional choice when necessary.

\begin{table*}[t]
\centering
\caption{Comparison with state-of-the-art methods on the R2R-CE val-unseen split. For fair comparison, zero-shot methods are grouped by evaluation setting. Here, ``waypoint-free'' indicates no reliance on an external learned waypoint proposal module.}
\label{tab:r2r_ce}
\resizebox{0.95\textwidth}{!}{%
\begin{tabular}{l>{\columncolor{gray!15}}c>{\columncolor{gray!15}}c>{\columncolor{gray!15}}c cccc}
\hline
\textbf{Method} & \textbf{Waypoint-free} & \textbf{Open-source Model} & \textbf{Model Size} & \textbf{NE}$\downarrow$ & \textbf{OSR}$\uparrow$ & \textbf{SR}$\uparrow$ & \textbf{SPL}$\uparrow$ \\
\hline
\multicolumn{8}{l}{\textit{Supervised Learning (full split)}} \\
\hline
Seq2Seq~\cite{beyond} & \cmark & -- & -- & 7.77 & 37.0 & 25.0 & 22.0 \\
MEE~\cite{mee} & \cmark & -- & -- & 6.82 & 44.6 & 35.9 & 32.3 \\
MLANet~\cite{he2025multilevel} & \cmark & -- & -- & 6.30 & 42.0 & 38.0 & 35.0 \\
VLN$\circlearrowright$BERT~\cite{vlnbert} & \xmark & -- & -- & 5.74 & 53.0 & 44.0 & 39.0 \\
ETPNav~\cite{etpnav} & \xmark & -- & -- & 5.15 & 65.0 & 57.0 & 49.0 \\
BEVBert~\cite{an2022bevbert} & \xmark & -- & -- & 4.57 & 67.0 & 59.0 & 50.0 \\
CLASH~\cite{wang2025clash} & \xmark & \cmark & 72B & \textbf{4.06} & \textbf{73.0} & \textbf{65.0} & \textbf{55.0} \\
\hline
\multicolumn{8}{l}{\textit{Zero-shot (full split)}} \\
\hline
COW-OWL~\cite{gadre2023cows} & \cmark & -- & -- & 8.72 & 5.9 & 3.4 & 1.6 \\
NavGPT-CE~\cite{zhou2024navgpt} & \xmark & \xmark & Undisclosed & 8.37 & 26.9 & 16.3 & 10.2 \\
CA-Nav~\cite{CANav} & \cmark & \xmark & Undisclosed & 7.58 & 48.0 & 25.3 & 10.8 \\
\textbf{P2DNav (Ours)} & \cmark & \cmark & 32B & \textbf{6.51} & \textbf{62.7} & \textbf{40.2} & \textbf{24.1} \\
\hline
\multicolumn{8}{l}{\textit{Zero-shot (100 episodes)}} \\
\hline
DiscussNav-CE~\cite{long2023discuss} & \xmark & \xmark & Undisclosed & 7.77 & 15.0 & 11.0 & 10.5 \\
MapGPT-CE~\cite{chen2024mapgpt} & \xmark & \xmark & Undisclosed & 8.16 & 21.0 & 7.0 & 5.0 \\
Open-Nav-Llama3.1~\cite{qiao2025open} & \xmark & \cmark & 70B & 7.25 & 23.0 & 16.0 & 12.9 \\
SmartWay~\cite{shi2025smartway} & \xmark & \xmark & Undisclosed & 7.01 & 51.0 & 29.0 & 22.5 \\
\textbf{P2DNav (Ours)} & \cmark & \cmark & 4B & \textbf{6.87} & \textbf{51.0} & \textbf{35.0} & 16.8 \\
\textbf{P2DNav (Ours)} & \cmark & \cmark & 8B & \textbf{6.95} & \textbf{59.0} & \textbf{38.0} & 16.6 \\
\textbf{P2DNav (Ours)} & \cmark & \cmark & 32B & \textbf{6.61} & \textbf{65.0} & \textbf{50.0} & \textbf{30.6} \\
\hline
\end{tabular}%
}
\end{table*}

\subsection{Reflective Reorientation Mechanism (RRM)}
\label{sec:rrm}

To handle these unreliable cases, we introduce the Reflective Reorientation Mechanism (RRM).
The key idea is that the downview stage should not only ground a local target, but also assess whether the selected observation provides sufficient evidence for reliable grounding.
Accordingly, the Stage-2 output is extended as
\begin{equation}
o_t^{(m)} \in \{p_t^{(m)},\ \texttt{abort}(r_t^{(m)})\},
\end{equation}
where $m$ denotes the reorientation round and $r_t^{(m)}$ denotes the reason for rejecting the current direction.
When the downview observation supports reliable grounding, the VLM returns a pixel-level target $p_t^{(m)}$ through the downview query $Q_{\mathrm{D}}$.
Otherwise, it returns an \texttt{abort} response with a concise reason, such as insufficient visual evidence, ambiguous traversable regions, or mismatch with the instruction.

Once \texttt{abort} is triggered, RRM records the rejected direction and its reason into the dialogue memory, and returns the decision process to panoramic direction selection.
The first stage then reconsiders the same panorama while excluding previously rejected directions:
\begin{equation}
d_t^{(m+1)} = Q_{\mathrm{P}}(I, P_t, \mathcal{M}_t, \mathcal{E}_t^{(m)}),
\end{equation}
where $\mathcal{E}_t^{(m)}=\{(d_t^{(i)}, r_t^{(i)})\}_{i=0}^{m}$ denotes the set of rejected directions and their corresponding reasons up to reorientation round $m$.
This exclusion mechanism prevents the model from repeatedly selecting the same unreliable direction, while the recorded reasons provide explicit feedback for choosing alternative directional candidates.
In practice, the rejected Stage-2 visual observation is not retained as a full image in memory; only the textual abort decision and reason are committed, which keeps the reorientation history useful without increasing visual-context cost.
The reorientation loop continues until a valid local target is produced, \texttt{STOP} is predicted, or the reorientation budget is exhausted.

RRM turns the two-stage P2D process into a closed-loop decision procedure.
Instead of passively executing every downview prediction, P2DNav uses local evidence from the second stage to revise earlier panoramic choices before action execution.
When local grounding is judged unreliable, the rejected direction and failure reason are recorded as explicit feedback for the next panoramic decision.
By integrating such feedback into SDM, RRM makes the hierarchical decision process reflective and revisable, improving navigation stability in cluttered or ambiguous scenes without requiring task-specific training.

\section{Experiments}
\label{experiment}

\subsection{Experimental Setup}

\textbf{Dataset:}
We evaluate P2DNav on the R2R-CE benchmark, a widely used continuous vision-and-language navigation dataset adapted from R2R to realistic 3D environments in the Habitat simulator~\cite{savva2019habitat}.
Unlike graph-based VLN, where the agent moves among predefined viewpoints, R2R-CE requires continuous navigation with low-level action execution.
This setting requires the agent to ground instructions into visual observations, infer movement directions, and generate executable local targets, making it suitable for evaluating zero-shot embodied navigation methods.
Following standard practice, we report results on the val-unseen split, which contains unseen environments and better reflects the agent's generalization ability.

\textbf{Evaluation metrics:}
We adopt standard VLN-CE metrics~\cite{anderson2018vision,wang2023res}, including Navigation Error (NE), Oracle Success Rate (OSR), Success Rate (SR), and Success weighted by Path Length (SPL). NE measures the distance between the agent's final position and the goal, where lower is better. OSR measures whether the trajectory reaches the success region at any point, regardless of the final stopping location, reflecting the agent's ability to approach the target. SR measures whether the agent successfully stops within the goal threshold, while SPL further accounts for path efficiency by penalizing unnecessarily long trajectories. Together, these metrics evaluate not only whether the agent reaches the goal, but also whether it can stop correctly and navigate robustly and efficiently.

\textbf{Implementation details:}
P2DNav is implemented in the Habitat simulator and performs zero-shot inference with Qwen3-VL~\cite{bai2025qwen3} as the navigator.
We use Qwen3-VL-32B as the default backbone model and evaluate other model scales in Sec.~\ref{sec:ablation}.
At each step, the agent constructs a stitched 360$^\circ$ panorama from six RGB views with 60$^\circ$ intervals for Stage-1 direction selection, and uses a direction-aligned downview RGB observation with a 15$^\circ$ downward tilt for Stage-2 local grounding. All RGB observations are resized to 224$\times$224 by default.
Stage 1 outputs \texttt{STOP} or \texttt{MOVE\_TO\_VIEW}, while Stage 2 outputs a pixel target or an \texttt{abort} response for RRM.
The predicted pixel is projected into world coordinates using aligned depth and executed by the low-level controller.
We set the SDM history window size to 1 and the RRM abort budget to the number of panoramic views.
All experiments are conducted on a single NVIDIA A800 GPU.
P2DNav runs in a fully zero-shot manner without task-specific VLN optimization or an additionally trained waypoint generator.

\begin{figure*}[t]
    \centering
    \includegraphics[width=0.8\textwidth]{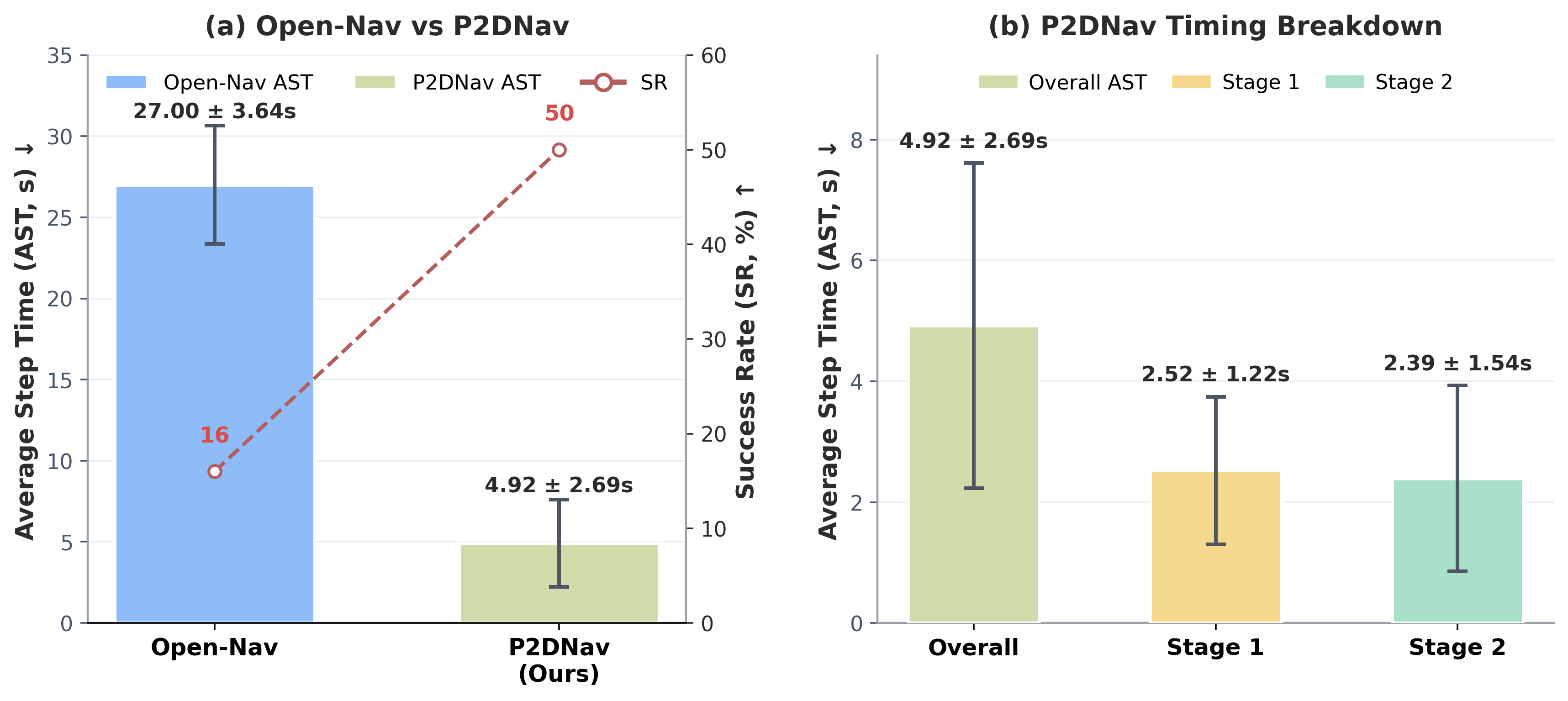}
    \caption{Inference efficiency analysis. (a) Comparison of average step time (AST) and success rate (SR) between Open-Nav and P2DNav. (b) Per-stage timing breakdown of P2DNav, including Stage 1 panoramic direction selection and Stage 2 downview local grounding.}
    \label{fig:time}
\end{figure*}

\subsection{Comparison with State-of-the-Art (SOTA)}

Table~\ref{tab:r2r_ce} compares P2DNav with supervised VLN methods and recent zero-shot baselines on the R2R-CE val-unseen split. 
For a fair comparison, we group zero-shot methods according to their evaluation setting, since some methods report results on the full val-unseen split while others evaluate on a 100-episode subset.

On the full val-unseen split, P2DNav achieves the best NE, OSR, and SR among zero-shot methods, obtaining 6.51 NE, 62.7 OSR, and 40.2 SR. 
Compared with the SOTA zero-shot waypoint-based baseline (NavGPT-CE), P2DNav improves SR from 16.3 to 40.2, corresponding to a relative gain of 146.6\%. 
Compared with the SOTA zero-shot waypoint-free baseline (CA-Nav), P2DNav improves SR from 25.3 to 40.2, yielding a relative gain of 58.9\%. 
These results show that explicitly decomposing panoramic direction selection and downview local grounding is effective for zero-shot continuous navigation. 

On the 100-episode evaluation setting following Open-Nav~\cite{qiao2025open}, P2DNav consistently outperforms prior zero-shot baselines across all metrics.
With the Qwen3-VL-32B backbone, P2DNav achieves 6.61 NE, 65.0 OSR, 50.0 SR, and 30.6 SPL, establishing the best performance in this setting.
Notably, even the 4B and 8B variants achieve competitive OSR and SR compared with previous zero-shot methods.
This suggests that the gains do not solely come from model scale, but from the proposed hierarchical, memory-aware, and reflective design.

\begin{table}[t]
\centering
\caption{Ablation study over the three core components of P2DNav: P2D, SDM, and RRM. The values in parentheses indicate the relative change compared with the plain base model.}
\label{tab:ablation_components}
\resizebox{0.5\textwidth}{!}{%
\begin{tabular}{ccc cccc}
\hline
\textbf{P2D} & \textbf{SDM} & \textbf{RRM} & \textbf{NE}$\downarrow$ & \textbf{OSR}$\uparrow$ & \textbf{SR}$\uparrow$ & \textbf{SPL}$\uparrow$ \\
\hline
\xmark & \xmark & \xmark & 7.15 & 35.0 & 26.0 & 12.7 \\
\cmark & \xmark & \xmark & 7.53 & 44.0 {\color{mygreen}(+9.0)} & 29.0 {\color{mygreen}(+3.0)} & 16.2 {\color{mygreen}(+3.5)} \\
\cmark & \cmark & \xmark & 6.93 & 54.0 {\color{mygreen}(+10.0)} & 42.0 {\color{mygreen}(+13.0)} & 24.1 {\color{mygreen}(+7.9)} \\
\cmark & \cmark & \cmark & \textbf{6.61} & \textbf{65.0} {\color{mygreen}(+11.0)} & \textbf{50.0} {\color{mygreen}(+8.0)} & \textbf{30.6} {\color{mygreen}(+6.5)} \\
\hline
\end{tabular}%
}
\end{table}

Compared with supervised methods, P2DNav still trails the SOTA trained models such as ETPNav, BEVBert, and CLASH in absolute SR and SPL. 
However, these methods rely on task-specific training and often use learned waypoint prediction or structured representation learning. 
In contrast, P2DNav operates in a fully zero-shot manner with an open-source multimodal model and does not depend on an additionally trained waypoint generator. 
Therefore, the comparison highlights the effectiveness of P2DNav as a training-free approach that substantially advances zero-shot VLN performance while narrowing the gap to supervised methods.

Fig.~\ref{fig:time} compares the inference efficiency of Open-Nav and P2DNav. As shown in Fig.~\ref{fig:time}(a), P2DNav reduces the average step time (AST) from 27.00$\pm$3.64\,s to 4.92$\pm$2.69\,s, which corresponds to an approximately 81.8\% reduction in per-step inference latency. At the same time, SR increases from 16\% to 50\%, showing that the proposed framework improves efficiency and effectiveness simultaneously, rather than trading one for the other.

The timing breakdown in Fig.~\ref{fig:time}(b) further shows that the latency of P2DNav comes from the two VLM inference stages. 
Stage 1 panoramic direction selection takes 2.52$\pm$1.22\,s, while Stage 2 downview local grounding takes 2.39$\pm$1.54\,s on average. 
The two stages have comparable runtime, indicating that neither stage dominates the overall latency.
The efficiency of P2DNav mainly comes from avoiding extra learned perception or waypoint proposal modules. 
Unlike methods that rely on separately trained waypoint generators or extra perception modules for object recognition and spatial description, such as FastSAM~\cite{zhao2023fast}, RAM~\cite{zhang2024recognize}, and SpatialBot~\cite{cai2025spatialbot}, P2DNav directly leverages the multimodal reasoning and grounding ability of the backbone VLM.
This design reduces system complexity and inference overhead, while our results suggest that VLMs can provide effective visual grounding and spatial reasoning capability for zero-shot navigation.

\begin{figure*}[t]
    \centering
    \includegraphics[width=0.98\textwidth]{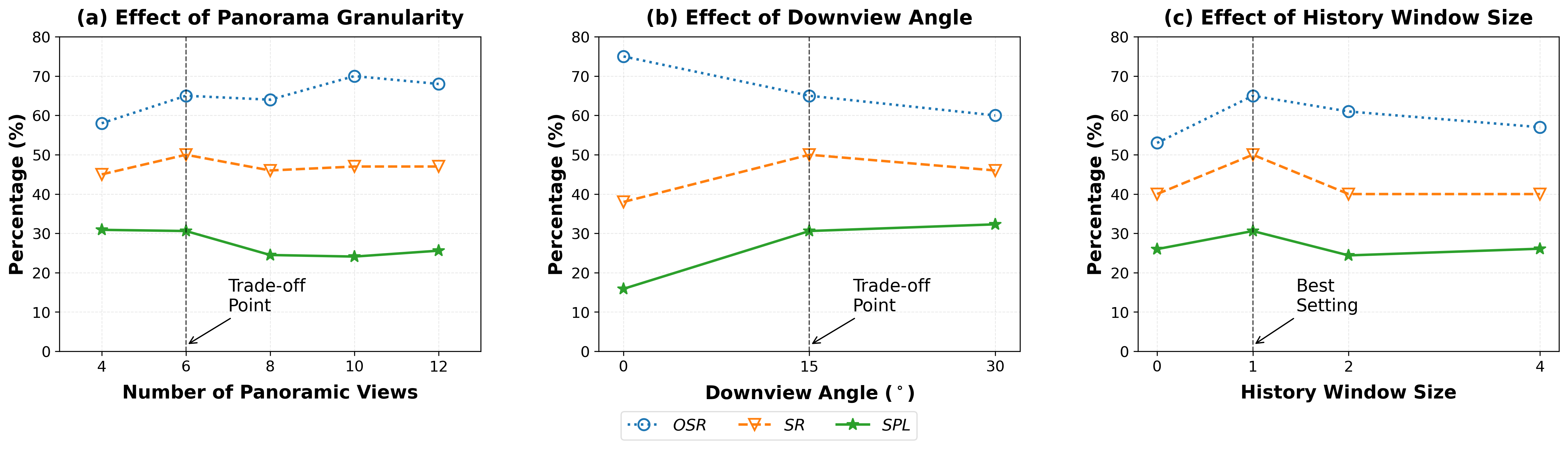}
    \caption{Design analysis of key choices in P2DNav. (a) Effect of panorama granularity. Using 6 panoramic views provides a favorable trade-off between directional coverage and decision complexity. (b) Effect of downview angle. A 15$^\circ$ downview angle achieves the best balance between global semantic awareness and local executable grounding. (c) Effect of the history window size in SDM. A window size of 1 yields the best overall performance, indicating that retaining a small amount of recent multimodal context is more effective than using fully compressed history or larger windows.}
    \label{fig:design_analysis}
\end{figure*}

\begin{table}[t]
\centering
\caption{Effect of model scale on P2DNav performance.}
\label{tab:model_scale}
\resizebox{0.42\textwidth}{!}{%
\begin{tabular}{lcccc}
\hline
\textbf{Model} & \textbf{NE}$\downarrow$ & \textbf{OSR}$\uparrow$ & \textbf{SR}$\uparrow$ & \textbf{SPL}$\uparrow$ \\
\hline
Qwen3-VL-2B  & 8.95 & 25.0 & 6.0  & 3.4 \\
Qwen3-VL-4B  & 6.87 & 51.0 & 35.0 & 16.8 \\
Qwen3-VL-8B  & 6.95 & 59.0 & 38.0 & 16.6 \\
Qwen3-VL-32B & \textbf{6.61} & \textbf{65.0} & \textbf{50.0} & \textbf{30.6} \\
\hline
\end{tabular}%
}
\end{table}

\subsection{Ablation Study}
\label{sec:ablation}

We conduct ablation experiments on the R2R-CE dataset to better understand the effectiveness of P2DNav. 
Our analysis focuses on six aspects: 
(1) the contribution of the three core components, 
(2) the influence of backbone model scale, 
(3) the effect of panorama granularity in Stage 1, 
(4) the impact of downview angle in Stage 2, 
(5) the role of SDM history window size and memory efficiency, and 
(6) the behavior of RRM in correcting unreliable local grounding.

\begin{figure}[t]
    \centering
    \includegraphics[width=0.5\textwidth]{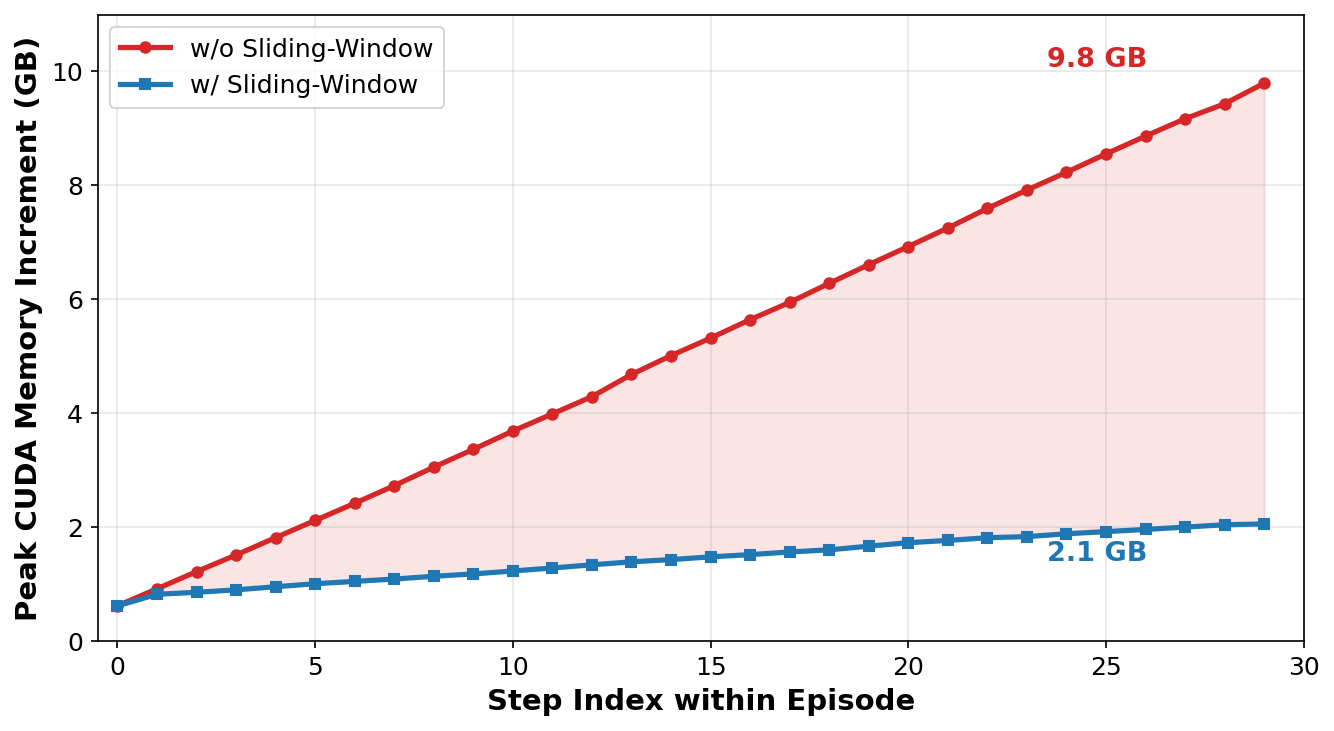}
    \caption{GPU memory increment with and without sliding-window memory management. By retaining only recent visual turns, SDM bounds visual-context growth and reduces the final peak CUDA memory increment from 9.8 GB to 2.1 GB.}
    \label{fig:vram}
\end{figure}

\begin{figure*}[t]
    \centering
    \includegraphics[width=0.8\textwidth]{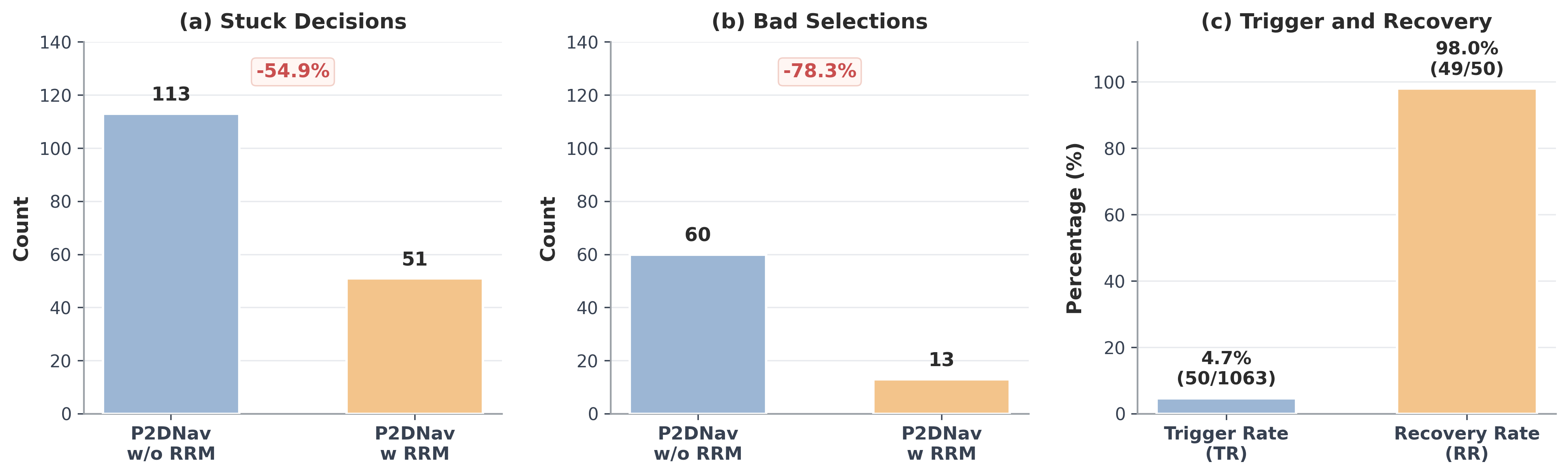}
    \caption{Analysis of the proposed Reflective Reorientation Mechanism (RRM). (a) RRM reduces the number of stuck decisions. (b) RRM also suppresses bad selections. (c) RRM is triggered only in a small fraction of steps (4.7\%), yet once triggered it is highly effective, achieving a recovery rate of 98.0\%.}
    \label{fig:RRM}
\end{figure*}

\begin{figure*}[t]
    \centering
    \includegraphics[width=0.95\textwidth]{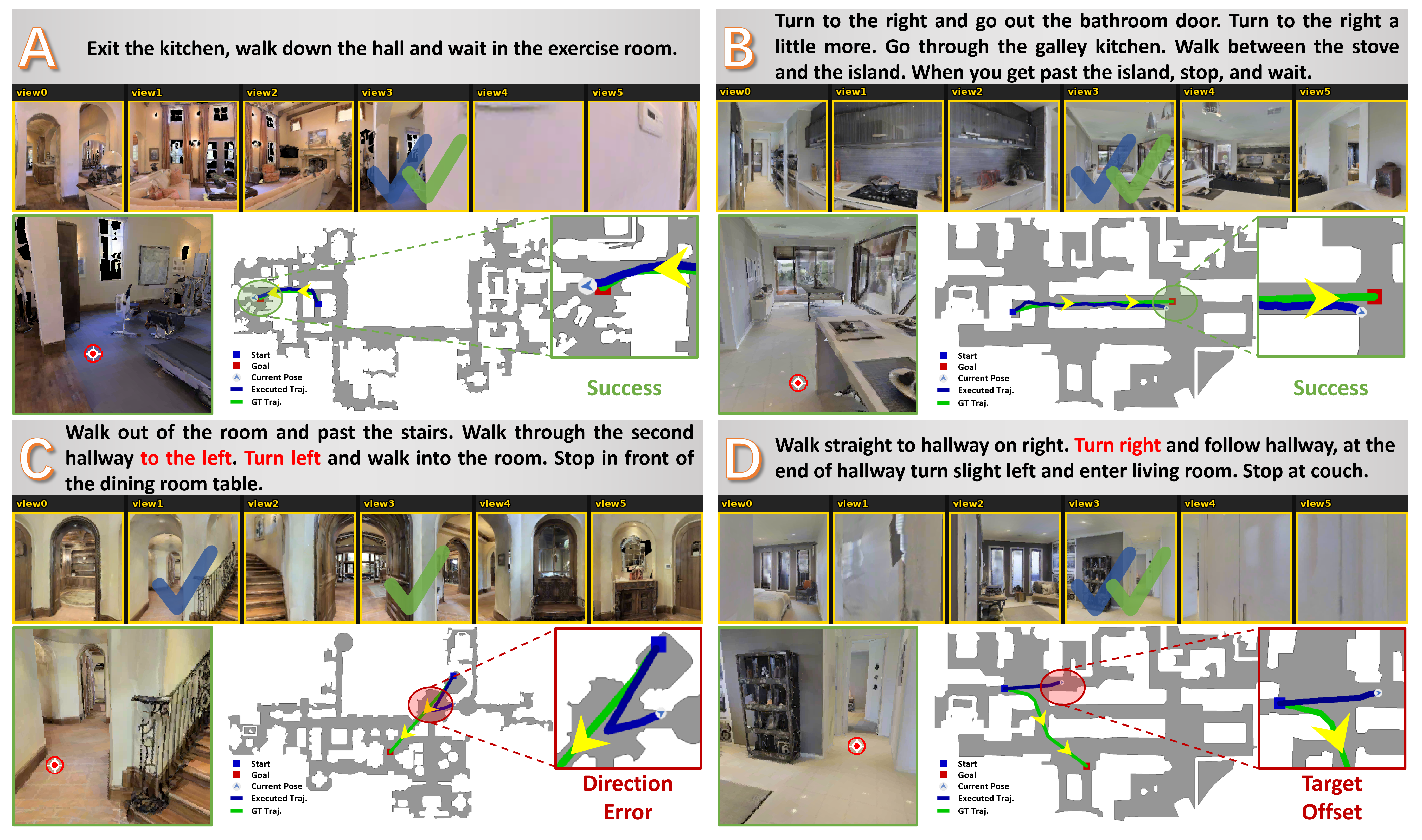}
    \caption{Qualitative case study of P2DNav. In each case, the top row shows the stitched panorama with six directional subviews, where \bluecheck\ denotes the predicted view and \greencheck\ denotes the ground-truth direction. \target\ in the 15$^\circ$ downview image indicates the predicted local target position for executable navigation. Cases A and B show successful examples with correct direction selection and accurate downview target prediction. Cases C and D show two representative failure modes: \textbf{Direction Error}, where panoramic direction selection chooses an incorrect subview, and \textbf{Target Offset}, where downview local grounding predicts an imprecise local target despite a generally correct direction.}
    \label{fig:fail}
\end{figure*}

\textbf{How much does each component contribute?}
We first evaluate the contribution of P2D, SDM, and RRM in Table~\ref{tab:ablation_components}. 
The plain base model follows the same two-stage decision protocol, but does not use the proposed P2D visual design. 
Specifically, in Stage 1, the six directional views are provided separately rather than stitched into a labeled panorama, making it harder to compare candidate directions in a unified spatial context. 
In Stage 2, the selected front-facing view is used instead of the 15$^\circ$ downview observation, reducing the local traversability cues available for pixel-level grounding. 

Adding the proposed P2D visual design increases OSR, SR, and SPL from 35.0, 26.0, and 12.7 to 44.0, 29.0, and 16.2, respectively. 
This improvement in success metrics indicates that panoramic stitching with view-index prompts and downview local grounding provide more effective visual representations for the two decision stages.

When SDM is introduced on top of P2D, OSR, SR, and SPL further increase to 54.0, 42.0, and 24.1, while NE decreases to 6.93. 
This shows that navigation benefits from preserving historical context across decision steps. 
Rather than relying on isolated single-turn prompts or repeatedly summarized memory, SDM maintains structured multi-turn navigation history while retaining recent visual evidence in a sliding window. 
Such memory provides richer context for long-horizon decision-making while keeping the multimodal history efficient and manageable.

Finally, enabling RRM yields the best overall performance, reaching 6.61 NE, 65.0 OSR, 50.0 SR, and 30.6 SPL. 
Compared with the plain base model, the full P2DNav improves OSR, SR, and SPL by 30.0, 24.0, and 17.9 points, respectively. 
The additional gains from RRM show the importance of reflective reconsideration before execution. 
By assessing whether the selected downview can support reliable grounding and returning uncertain cases to panoramic direction selection, RRM prevents unreliable local predictions from being directly translated into actions.

These results show that the three components are complementary: P2D provides stage-specific visual representations, SDM supplies efficient and comprehensive historical memory, and RRM introduces a reflective correction loop for unreliable local grounding.

\textbf{How does the backbone scale affect P2DNav?}
We then analyze the effect of model scale in Table~\ref{tab:model_scale}. 
Overall, performance improves substantially as the backbone becomes larger, with Qwen3-VL-32B achieving the best results across all metrics.
In particular, the 32B model reaches 6.61 NE, 65.0 OSR, 50.0 SR, and 30.6 SPL, showing that stronger multimodal reasoning capacity is beneficial for P2DNav.
However, the gains do not solely come from model scaling. 
Even Qwen3-VL-4B already obtains 51.0 OSR and 35.0 SR, matching or surpassing several prior zero-shot baselines on these two metrics.
In contrast, Qwen3-VL-2B performs poorly, suggesting that small models still lack sufficient multimodal reasoning and grounding ability for reliable continuous zero-shot navigation.

\textbf{What is the best panoramic granularity for Stage 1?}
Fig.~\ref{fig:design_analysis}(a) studies the effect of panorama granularity. 
Using too few views makes directional partitioning coarse and weakens global direction selection, while using too many views introduces redundant visual content and increases decision difficulty.
Empirically, $K=6$ provides the best balance between directional coverage and reasoning complexity, achieving the most balanced performance across metrics.
This supports our choice of using six directional subviews in Stage 1.

\textbf{What downview angle is suitable for Stage 2?}
Fig.~\ref{fig:design_analysis}(b) analyzes the effect of downview angle. 
Without downward tilting, the model can still achieve high OSR, but SR and SPL remain limited, indicating that local grounding is less stable.
A moderate downview angle of 15$^\circ$ significantly improves SR and SPL by exposing more nearby floor regions and local geometric cues.
When the angle is increased to 30$^\circ$, SPL slightly improves, but OSR and SR decrease, likely because excessive downward tilting weakens directional consistency with the selected panoramic view.
Therefore, 15$^\circ$ offers the best trade-off between local geometric visibility and directional consistency.

\textbf{How much history should SDM retain?}
Fig.~\ref{fig:design_analysis}(c) examines the history window size in SDM. 
When the window size is 0, all historical visual turns are converted into text-only memory, leading to clear performance degradation.
Increasing the window size to 1 significantly improves OSR, SR, and SPL, showing that retaining recent multimodal context is important for stable decision-making.
However, further increasing the window size to 2 or 4 does not bring additional gains and may reduce performance on several metrics.
This suggests that SDM benefits from keeping the most recent visual evidence, while overly large windows may introduce redundant historical information and make reasoning more difficult.
Overall, a window size of 1 is the optimal setting in our experiments.

Beyond navigation performance, SDM also improves memory efficiency.
As shown in Fig.~\ref{fig:vram}, preserving all historical visual turns causes peak CUDA memory to grow almost linearly as the episode proceeds.
By contrast, SDM bounds visual-context growth by retaining only recent visual turns, reducing the final peak CUDA memory increment from 9.8 GB to 2.1 GB.
This confirms that the sliding-window design provides efficient multimodal memory without unbounded visual-context growth.

\textbf{When and how does RRM help?}
We further analyze the behavior of RRM in Fig.~\ref{fig:RRM}. 
First, RRM reduces obviously failure-prone local decisions.
As shown in Fig.~\ref{fig:RRM}(a), the number of \emph{stuck decisions} decreases from 113 to 51 after enabling RRM, corresponding to a 54.9\% reduction. 
A stuck decision is defined by $\texttt{local\_steps} \leq 1$, where the low-level controller executes at most one action and the agent almost fails to move.
Similarly, Fig.~\ref{fig:RRM}(b) shows that \emph{bad selections}, defined by local targets with $\texttt{pixel\_depth} < 0.25$m, drop from 60 to 13, corresponding to a 78.3\% reduction. 
Such shallow targets often correspond to nearby obstacles or invalid regions.
These results indicate that RRM helps the agent avoid clearly unreasonable local grounding decisions before execution.
Second, RRM is lightweight in practice.
As shown in Fig.~\ref{fig:RRM}(c), its trigger rate (TR) is only 4.7\%, meaning that reflective reconsideration is invoked only in a small fraction of high-risk cases.
However, the recovery rate (RR) reaches 98.0\% (49/50), showing that once RRM intervenes, it almost always prevents the two types of failure-prone local decisions discussed above.
Thus, RRM is neither a high-frequency interruption mechanism nor a passive fallback.
Instead, it acts as a selective closed-loop correction module that intervenes only when necessary and substantially improves the reliability of local decision-making.

\subsection{Qualitative Case Study}
\label{sec:failure}

To better understand the behavior and remaining limitations of P2DNav, we provide a qualitative case study in Fig.~\ref{fig:fail}.
Cases A and B show successful examples, where P2DNav selects the correct panoramic direction and predicts an accurate downview target for executable navigation.
These examples indicate that the two-stage design can effectively align coarse directional reasoning with fine-grained local grounding when the relevant direction is visible and the downview provides reliable traversable cues.

Cases C and D reveal two representative failure modes: \emph{Direction Error} and \emph{Target Offset}.
Direction Error occurs when panoramic direction selection chooses an incorrect subview, causing local grounding to follow an erroneous global direction.
As shown in Case C, the agent follows the instruction and turns left, but this local instruction cue is not fully aligned with the global reference trajectory, leading to failure.
Target Offset occurs when the selected direction is generally correct, but the downview stage predicts an imprecise pixel-level target.
As shown in Case D, the agent moves past the intended intersection; when it later needs to turn right, the right-facing downview no longer contains a reliable traversable region, causing grounding to be aborted and the turning opportunity to be missed. These cases suggest that P2DNav performs well when both direction selection and downview grounding are reliable, while its remaining challenges lie in coarse directional selection and fine-grained local target prediction.
Panoramic direction selection may be affected by instruction ambiguity, trajectory-instruction mismatch, or visually similar candidate views, while downview grounding still requires more precise and timely target prediction near intersections, doorways, and turning points.
Therefore, improving coarse direction reliability and fine-grained grounding accuracy remains an important direction for future work.

\section{Conclusions}

In this paper, we presented P2DNav, a zero-shot framework for continuous vision-and-language navigation. Instead of treating zero-shot VLN as a one-step action or waypoint prediction problem, P2DNav reformulates it as a hierarchical, memory-aware, and revisable decision process through three complementary modules: Panorama-to-Downview (P2D), Sliding-Window Dialogue Memory (SDM), and Reflective Reorientation Mechanism (RRM). Specifically, P2D separates panoramic direction selection from downview local grounding, SDM maintains efficient multi-turn historical context, and RRM enables reorientation when local grounding is unreliable. Together, these modules improve the conversion from high-level semantic intent to low-level executable navigation decisions.
Experiments on R2R-CE demonstrate that P2DNav achieves state-of-the-art performance among zero-shot methods and outperforms prior waypoint-based and waypoint-free baselines. Further ablations and behavior analyses verify the complementary contributions of P2D, SDM, and RRM. These results suggest that zero-shot VLN performance is limited not only by instruction understanding, but also by how reliably semantic intent is grounded into executable local targets. We hope this work can inspire future foundation-model-driven navigation systems to better integrate hierarchical reasoning, efficient memory, and reflective correction in continuous environments.




 

\bibliographystyle{IEEEtran}
\bibliography{ref}


 





\end{document}